%% file: main.tex
\documentclass[letterpaper]{article} 
\input{packages}

\input{define}

\title{Debiased Fine-Tuning for Vision-Language Models by Prompt Regularization}

\author{
 \textbf{Beier Zhu}\textsuperscript{\rm 1}, \textbf{Yulei Niu}\textsuperscript{\rm 2}\thanks{Corresponding author. Work started when at NTU.}, \textbf{Saeil Lee}\textsuperscript{\rm 3}, \textbf{Minhoe Hur}\textsuperscript{\rm 4}, \textbf{Hanwang Zhang}\textsuperscript{\rm 1}}

\affiliations {
    \textsuperscript{\rm 1} Nanyang Technological University\\
    \textsuperscript{\rm 2} Columbia University \\
    \textsuperscript{\rm 3} HMGICS AIR Center \\
    \textsuperscript{\rm 4} AIRS Company, Hyundai Motor Group \\
    beier002@e.ntu.edu.sg, yn.yuleiniu@gmail.com, saeil.lee@hmgics.com, minhoe.hur@hyundai.com, hanwangzhang@ntu.edu.sg
}

\begin{document}

\maketitle

\input{sections/0-abstract}
\input{sections/1-intro}
\input{sections/2-related}
\input{sections/3-method}
\input{sections/4-experiments}

\input{sections/5-conclusion}
\input{sections/6-ack}

\bibliography{aaai23}

\end{document}

%% file: packages.tex
\usepackage{aaai23}  
\usepackage{times}  
\usepackage{helvet}  
\usepackage{courier}  
\usepackage[hyphens]{url}  
\usepackage{graphicx} 
\usepackage{natbib}  
\usepackage{caption} 
\usepackage{algorithm}
\usepackage{algorithmic}
\usepackage{newfloat}
\usepackage{listings}
\usepackage{bibentry}
\usepackage{color, colortbl}

\usepackage{tikz}
\usepackage{comment}
\usepackage{amsmath,amssymb} 
\usepackage{bm}
\usepackage{mathtools} 
\usepackage{dsfont} 
\usepackage{bbm}
\usepackage{booktabs} 
\usepackage{diagbox} 
\usepackage{nccmath} 
\usepackage{arydshln} 
\usepackage{tablefootnote}
\usepackage{amssymb}
\usepackage{enumitem} 
\usepackage{subfigure}
\usepackage{amsmath}
\usepackage{xcolor}
\usepackage{multirow}
\usepackage{tabu}
\usepackage{soul}
\usepackage{pifont} 
\usepackage{physics}

%% file: define.tex
\DeclareCaptionStyle{ruled}{labelfont=normalfont,labelsep=colon,strut=off} 
\floatstyle{ruled}
\newfloat{listing}{tb}{lst}{}
\floatname{listing}{Listing}
\newlength\savewidth

\newcolumntype{x}[1]{>{\centering\arraybackslash}p{#1pt}}
\newcolumntype{y}[1]{>{\raggedright\arraybackslash}p{#1pt}}
\newcolumntype{z}[1]{>{\raggedleft\arraybackslash}p{#1pt}}

\definecolor{ABOVE}{HTML}{39b54a}  
\definecolor{BELOW}{HTML}{FF0000}  

\newcolumntype{Q}{@{}c@{}}

\definecolor{orange}{HTML}{FF7F00}
\definecolor{col1}{HTML}{b4ce08}
\definecolor{col2}{HTML}{407934}
\definecolor{col3}{HTML}{353B3D}
\definecolor{airforceblue}{rgb}{0.36, 0.54, 0.66}
\definecolor{bleudefrance}{rgb}{0.19, 0.55, 0.91}
\definecolor{bluemunsell}{rgb}{0.0, 0.5, 0.69}
\definecolor{darkpastelgreen}{rgb}{0.01, 0.75, 0.24}
\definecolor{armygreen}{rgb}{0.29, 0.33, 0.13}
\definecolor{cadmiumgreen}{rgb}{0.0, 0.42, 0.24}
\definecolor{darkspringgreen}{rgb}{0.09, 0.45, 0.27}
\definecolor{ferngreen}{rgb}{0.31, 0.47, 0.26}
\definecolor{forestgreen(web)}{rgb}{0.13, 0.55, 0.13}
\definecolor{grannysmithapple}{rgb}{0.66, 0.89, 0.63}
\definecolor{green(html/cssgreen)}{rgb}{0.0, 0.5, 0.0}
\definecolor{mygray}{gray}{0.9}

\definecolor{col4}{HTML}{4B5232}

\newcommand{\textcite}{\cite}

\usepackage{array}
\newcolumntype{P}[1]{>{\centering\arraybackslash}p{#1}}

\newcommand{\eg}{\textit{e.g.}}
\newcommand{\ie}{\textit{i.e.}}

\newcommand{\etc}{\textit{etc}}

\newcommand{\Lkd}{\mathcal{L}_\text{kd}}
\newcommand{\Lce}{\mathcal{L}_\text{ce}}
\newcommand{\Lkl}{\mathcal{L}_\text{kl}}
\newcommand{\LProReg}{\mathcal{L}_\text{ProReg}}
\newcommand{\yzs}{\mathbf{y}^\text{zs}}
\newcommand{\zt}{\mathbf{z}_t}

\newcommand{\mask}{\texttt{[MASK]}}
\newcommand{\class}{\texttt{[CLASS]}}

\definecolor{tabhighlight}{HTML}{e5e5e5}

%% file: sections/0-abstract.tex
\begin{abstract}
We present a new paradigm for fine-tuning large-scale vision-language pre-trained models on downstream task, dubbed Prompt Regularization (ProReg). Different from traditional fine-tuning which easily overfits to the downstream task data, ProReg uses the prediction by prompting the pretrained model to regularize the fine-tuning. The motivation is: by prompting the large model ``a photo of a \class'', the fill-in answer is only dependent on the pretraining encyclopedic knowledge while independent of the task data distribution, which is usually biased. Specifically, given a training sample prediction during fine-tuning, we first calculate its Kullback-Leibler loss of the prompt prediction and Cross-Entropy loss of the ground-truth label, and then combine them with a proposed sample-wise adaptive trade-off weight, which automatically adjusts the transfer between the pretrained and downstream domains. On various out-of-distribution benchmarks, we show the consistently strong performance of ProReg compared with conventional fine-tuning, zero-shot prompt, prompt tuning, and other state-of-the-art methods. 
\end{abstract}

%% file: sections/1-intro.tex
\section{Introduction}\label{sec:1}
\input{images/Figure1}

Please think about it: when you want to train a vision model for a task, what is the first thing first? Most likely, you will download an off-the-shelf foundation model~\cite{bommasani2021opportunities}, \eg, ResNet~\cite{resnet} or CLIP~\cite{clip}, pretrained on a large-scale dataset such as ImageNet~\cite{imagenet} or
image-text pairs collected from the Internet~\cite{clip}; then, remove its classifier head, plug in your own task-specific head to the penultimate layer, and finally fine-tune on your own task data. Such ``pretrain, fine-tune'' paradigm has become a nearly ubiquitous standard for CV community---from classification to generation~\cite{stylegan2,biggan}, regions to pixels~\cite{maskrcnn}, and single modality to multi-modality~\cite{UpDn,ViLBERT}. The underlying empirical principle is that the pretrained model as initialization plays a role in regularization which reduces the variance of the fine-tuning model~\cite{sutskever2013importance}. 

Despite the beneficial regularization, the pretraining knowledge has a negative impact, especially when the downstream task data is limited or biased~\cite{negtransfer,yang2021causal}: the early exposed encyclopedic or generic features from the pretrained model may mislead the fine-tuning to focus on the task-unrelated attributes, resulting in a biased fine-tuned model. Figure~\ref{fig:figure1} shows three types of biases: Figure~\ref{fig:figure1}(a) contextual bias: images of training and test set contain distinct backgrounds, \eg, if most of the training ``bird'' images are in ``sky'' background, the fine-tuning will misclassify ``bird on ground'' as ``dog''; Figure~\ref{fig:figure1}(c) image style 
gap: test image style is unseen during training, \eg, if training images are from art painting, cartoon and real domain, the fine-tuned model is able to classify the in-distribution art painting ``dog'' but is confused when testing on a sketch ``dog'';  Figure~\ref{fig:figure1}(e) language bias: for every question type, train and test sets have different prior distributions of answers, \eg, if most VQA training ``bananas'' images are ``yellow'', it mistakes the answer ``yellow'' for question ``what color are the bananas?'', given an image of green bananas.  

Recently, NLP community presents a tuning-free paradigm called ``pretrained, prompt, predict''~\cite{promptsurvey} and it has been quickly migrated to CV tasks~\cite{coop,frozen,cpt} by using a pretrained multi-modal model~\cite{vilt,clip,vinvl}. For example, image classification can be cast into the cloze prompt: ``a photo of a \class'', where the prediction can be found as the class word whose fill-in sentence is most similar to the query image. The similarity can be calculated directly from the pretrained model in a zero-shot learning fashion. As prompt is a rule-based query that has nothing to do with the downstream statistics, the prediction is expected to be independent to downstream domain and faithfully respects the pretrained knowledge. 

Yet, relying too much on the general knowledge also hurts the domain-specific generalization. For example, as shown in Figure~\ref{fig:figure1}(b), although prompt can correctly focus on the foreground object, it is less discriminative to distinguish between ``rat'' and ``cat'' in domain-specific animal images; 
in Figure~\ref{fig:figure1}(d), prompt's prediction confidence is not as discriminative as that of fine-tuning; in Figure~\ref{fig:figure1}(f), prompt-based VQA is too general to perform the downstream task of counting ``bananas''. To this end, ``prompt tuning'' is proposed to fine-tune the token embeddings in a prompt using the task data~\cite{promptsurvey,liu2021gpt,han2021ptr,coop}. For example, the prefix ``a photo of a'' before the cloze blank \class can be replaced with a set of tunable parameters. Prompt tuning is essentially fine-tuning with fixed backbone and tunable head (\ie, the prompt prefix). Therefore, it still inherits the above drawbacks of the biased fine-tuning.

In this paper, we present a new fine-tuning paradigm, called \emph{Prompt Regularization} (ProReg), for a \emph{just-right} knowledge transfer from pretrained model to fine-tuned model. As expected, ProReg can fine-tune the resultant model, neither biased towards the pretrained knowledge nor towards the downstream knowledge. 
We formulate the downstream knowledge as the ground-truth annotations in downstream tasks, and represent the pretrained knowledge with the ``soft" label of the downstream data generated by prompting the pre-trained model.
We then proposed the ProReg loss to enable learning from both of the knowledge.
It is worth noting that different from traditional knowledge distillation that using a constant weight $\lambda \in (0,1)$ to trade-off the contribution of the knowledge: $\Lkd=(1-\lambda)\cdot \Lce+\lambda\cdot \Lkl$, 
we propose a sample-wise adaptive weight to achieve a good trade-off between the two knowledge (Section~\ref{sec:proreg}). The proposed weight inspects whether the task-specific knowledge or the pre-trained general knowledge dominates the optimization process for each fine-tuning sample, which indeed requires a different ratio of the two knowledge types. We show that the estimation of the ratio evolves during the training process and  can be automatically calculated on-the-fly.

We implement ProReg on top of two off-the-shelf large-scale pretrained models: CLIP~\cite{clip} and ViLT~\cite{vilt}, which demonstrates that ProReg is applicable to different vision-language models that adopt masked language modeling or contrastive learning as pretraining tasks. 
We conduct extensive evaluations for ProReg on various out-of-distribution benchmarks, including BAR~\cite{LfF}, NICO~\cite{nico}, PACS~\cite{PACS} and DomainNet~\cite{DomainNet} for image classification tasks and  VQA-CP~\cite{VQACP-GVQA} for visual question answering tasks. 
We demonstrate that: 1) ProReg consistently outperforms zero-shot prompt, conventional fine-tuning, and prompt tuning on all the datasets, 2) ProReg achieves compelling performance in both out-of-distribution and in-distribution settings. Thus, readers can feel free to use ProReg regardless of the training and testing distribution discrepancy.

%% file: images/Figure1.tex
\begin{figure*}[t]
\centering
\includegraphics[width=0.95\textwidth]{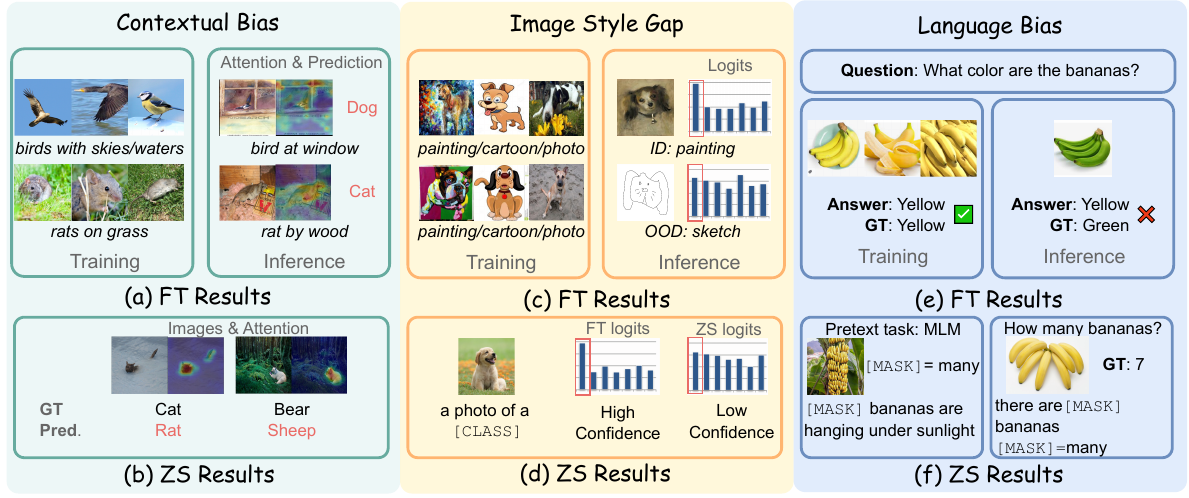}
\caption{
Examples of the contextual bias~\cite{nico,LfF}, image style bias~\cite{PACS} and language bias~\cite{VQAv2,VQACP-GVQA} caused by only relying on fine-tuning. (a\&c\&e): If the downstream data is biased, fine-tuning is biased. (b\&d\&f): Pretraining knowledge is too general to perform domain-specific downstream tasks.}
\label{fig:figure1}
\end{figure*}

%% file: sections/2-related.tex
\section{Related Work}
\noindent\textbf{Vision-Language Models~(VLM).}
Most existing VLMs use 1) Masked language modeling~\cite{vilt,visualbert,ViLBERT}, 2) Image text pairing~\cite{lxmert,PixelBERT}, and 3) Contrastive learning~\cite{clip} as their pretraining objectives. 
Recently there is a line of adapting the existing VLMs to the downstream tasks. Conventional fine-tuning paradigm adds an additional classifier on top of the visual backbone (Linear Probe~\cite{clip}) or additional feature adapter, (CLIP-Adapter~\cite{gao2021clip}). Prompt-based learning that tunes the prompt to maximize the ground-truth token has gained its popularity, \eg, CoOp~\cite{coop} and CoCoOp~\cite{zhou2022cocoop}. ProGrad~\cite{ProGrad} bridges generalization gap by matching the gradient of prompt to the general knowledge. ProDA~\cite{ProDA} introduces prompt distribution learning to adapts VLMs to downstream classification tasks. 
As discussed in Section~\ref{sec:1}, both of these two fine-tune paradigms may result in biased downstream models, in this work, we aims to fine-tune a debiased VLM for downstream tasks.


\noindent\textbf{OOD Generalization}.
In real world, test distribution may shift from the training distribution, such as domain generalization~\cite{ben2007analysis,tzeng2017adversarial}, long-tailed classification~\cite{menon2020long,tang2020long}, contextual bias~\cite{LfF,nico}, and language bias~\cite{SCR,advreg,rubi,CF-VQA}. WiSE~\cite{WiSE} shares the same goal to improve OOD performance, which ensembles the weights of zero-shot and fine-tuned models. However, it requires the fine-tuned model and the pre-trained model to have the same architecture.
Differently, our ProReg is free of this requirement and allows the modification of architecture.
\input{images/ProReg}

%% file: images/ProReg.tex
\begin{figure*}
    \centering
    \includegraphics[width=1\textwidth]{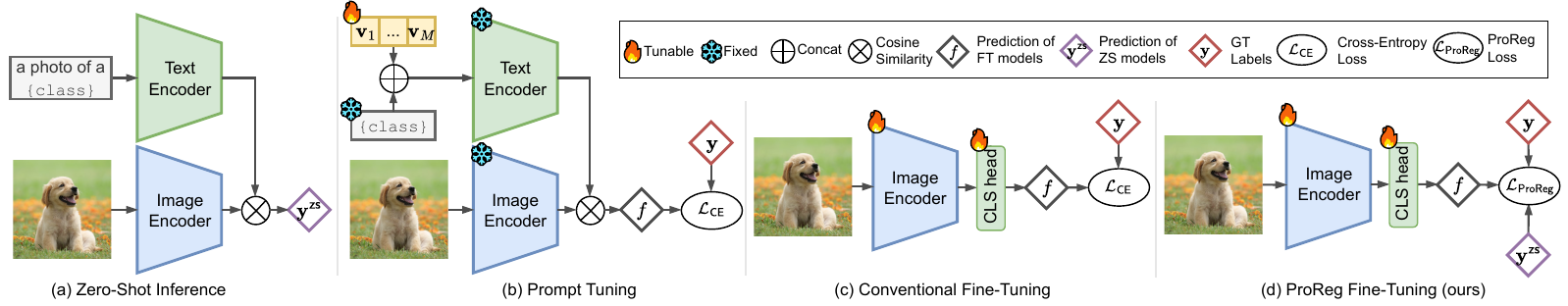}
    \caption{Comparisons of frameworks based on CLIP models for image classification.
    (a) Zero-shot Prompt Inference, \ie, ``pretrain, prompt'', generates the classification weights by a text encoder thought prompting.
    (b) Prompt Tuning, \ie, ``pretrain, prompt, fine-tune'',  learns continuous prompt by minimizing the cross-entropy loss.
    (c) Conventional Fine-tuning, (\ie, ``pre-train, fine-tune''), adds classification head on top of the visual backbone and is optimized by cross-entropy loss. The classification head can be randomly initialized or initialized by the text encoder thought prompting. (d) ProReg fine-tuning. Different from conventional fine-tuning, the supervisions come from two sources: the ground-truth target from the downstream domain, and the zero-shot prompt prediction from the pretraining domain. The classification head of ProReg is initialized by prompting.}
    \label{fig:proreg}
\end{figure*}

%% file: sections/3-method.tex
\section{The ProReg Pipeline}

\subsection{Pre-trained Model}
In this paper, we adopt two state-of-the-art VLM models as our backbones, \ie, Constrastive Language-Image Pre-training model (CLIP~\cite{clip} for image classification tasks and Vision-and-Language Transformer (ViLT~\cite{vilt})) for VQA tasks. 

CLIP collects large amounts of images with natural language description and is trained in a contrastive manner.
The associated image and text pairs are regarded as positive pairs, while the mis-matched image and text pair are regarded as negative ones.
The contrastive loss aims to maximize the cosine similarity of the positive pairs in the batch while minimize the cosine similarity of the  negative pairs. 

ViLT consists of stacked blocks that include a multi-headed self-attention and a multi-layer perceptron layer.
It is pretrained with two pretext tasks: Image Text Matching (ITM) and Masked Language Modeling (MLM).
For ITM, ViLT selects half of the sentences and replaces each of them with a mismatched sentence.  
Then, a binary classifier is added to predict whether the image and the sentence match with each other. 
MLM first randomly replace $15\%$ of input text tokens with masked tokens, \eg, \texttt{[MASK]}, and then the model is optimized to predict the masked tokens, given other non-masked tokens and image feature. 

For classification tasks, CLIP based models show superior performance over the ViLT one, thus we only report the CLIP results in the main paper while the image classification results of ViLT can be found in Appendix.
For VQA tasks, the input text varies sample-wisely, it is difficult for CLIP model to infer and optimize. As a result, we only implement the ProReg on ViLT models.
Note that our approach is applicable to broader vision-language models whose pretrained pretext tasks include ITM and MLM objectives or contrastive objective.

\subsection{Prompt-based Zero-shot Inference}\label{sec:promptdesign}
Proper prompt can adapt the pre-trained model to downstream tasks without fine-tuning. 
In this section, we illustrate how CLIP performs zero-shot inference for image classification tasks and how ViLT can leverage carefully designed prompt to perform zero-shot prediction for VQA tasks.
\input{tables/prompt_example}

\noindent\textbf{Image Classification}.
We formulate image classification as an image-text matching problem where the text is designed by a simple static prompt template. Take the action recognition dataset BAR~\cite{LfF} in Figure~\ref{fig:proreg}(a) as an example. We first compute the text embedding $\mathbf{w}_i$ of ``a person is \class.'', where the ``\class'' is replaced by $i$-th class name. Then the probability distribution of the  image feature $\mathbf{x}$ over $K$ classes is given by:

\begin{equation}\label{eq:clipzs}
    f(y=i|\mathbf{x};\theta)=\frac{\exp(<\mathbf{x},\mathbf{w}_i>\!/\tau)}{\sum_{j=1}^K\exp(<\mathbf{x},\mathbf{w}_j>\!/\tau)}
\end{equation}
where $\tau$ is the temperature learned by CLIP.

\noindent\textbf{Visual Question Answering (VQA)}.
The design of prompts for VQA task is less straightforward because the text inputs in VQA (\ie, questions) and pretraining tasks (\ie, captions) have different forms. 
Notice that VQA dataset consists of two types of questions: open-ended questions (\eg, ``what color is the cake?'') and closed-ended questions (\eg, ``any brown apples in the picture?''). We convert the question to a statement and name such design as \textbf{Question-to-Statement (Q2S)} prompt. Table~\ref{tab:promptexample} shows some prompt examples. Specifically, we use ITM for closed-ended questions and MLM for open-ended questions. For example, for a closed-ended question like ``any brown apples in the picture?'', the prompt statement is generated as ``some brown apples in the picture.''. The text embedding together with the visual feature are fed into ITM head to predict whether they match with each other. A high match score corresponds to the answer ``yes'' and a low score to ``no''.
For an open-ended question like ``what color is the cake?'', the prompt is generated as ``the color of the cake is \mask.''. Similar to Eq.~\eqref{eq:clipzs}, we search the candidate answer set to find the one has the highest probability.

\subsection{Sample-wise Adaptive Trade-off Weight}\label{sec:proreg}
In this section, we illustrate how to implement ProReg for CLIP models, the ProReg for ViLT models can be implemented analogously or refer to Appendix.
Figure~\ref{fig:proreg} (b) and (c) show the comparison of current fine-tuning works. 
Figure~\ref{fig:proreg} (b) shows the pipeline of ``pretrain, prompt, fine-tune'' where the classification head is generated by the optimizing the continuous prompt.
We name the ``pretrain, fine-tune'' paradigm in Figure~\ref{fig:proreg}(c) as ``\textit{Conventional Fine-tuning}'', which adds a classification head for prediction. The classification head can be randomly initialized or initialized by feeding the hand-craft prompt to the text encoder. 
The common characteristic is that both of the two paradigms only use the ground-truth labels as supervision.
As discussed in Section~\ref{sec:1}, both of these two fine-tune paradigms may result in biased downstream models, especially when the task data is limited and biased~\cite{ifsl,negtransfer,yang2021causal}.

We present a new fine-tuning paradigm, dubbed \emph{Prompt Regularization} (ProReg), 
to transfer the knowledge from the pretraining domain to the downstream domain.
ProReg considers two types of supervisions for optimization: 
task-specific downstream knowledge and task-agnostic general knowledge.

\noindent \textbf{Task-specific downstream knowledge}. The ground-truth labels serve as the task-specific knowledge and allows the model $f(;\theta)$ to be adapted to downstream task. The cross-entropy $\Lce$ of an image $\mathbf{x}$ is obtained as:
\begin{equation}
   \Lce(\theta) = -\sum_i \mathbf{y}_i \log f_i(\mathbf{x};\theta),
\end{equation}
where $\mathbf{y}$ denotes the one-hot ground-truth vector, and the subscript $i$ denotes the $i$-th label; $f(;\theta)$ is the classification model initialized by pretrained model. 

\noindent \textbf{Task-agnostic general knowledge}. Compared to large-scale datasets for pre-training, the task-specific data for downstream task may be limited or even biased. As a result, only taking the ground-truth annotations as supervision may lead to biased fine-tuning. In order to achieve debiased fine-tuning, we also use task-agnostic pre-training knowledge as regularization. We take the zero-shot prompt prediction of the pre-trained model $\yzs$ as the regularization knowledge, and use kullback-leibler divergence ($\Lkl$) between the fine-tuned model prediction and the zero-shot prompt prediction as the regularization term:

\begin{equation}\label{eq:klloss}
    \Lkl(\theta) = - \sum_i \yzs_i\log \frac{f_i(\mathbf{x};\theta)}{\yzs_i} ,
\end{equation}

To fine-tune the model neither biased towards the downstream domain nor the pre-trained knowledge domain, a straightforward intuition is knowledge distillation (KD)~\cite{hinton2015distilling}, to combine the two terms with a constant weight $\lambda \in (0,1)$:
\begin{equation}\label{eq:totalloss1}
\Lkd=(1-\lambda)\cdot\Lce +\lambda\cdot \Lkl,\\
\end{equation}
where we omit $\theta$ for simplicity. 
However, this simple solution overlooks that the trade-off of the two knowledge varies sample-wisely and evolves during training. 
In the next part, we will analyze the trade-off from the view of task-agnostic knowledge decomposition and propose a dynamic weight to balance the two loss terms for each training sample.

\noindent\textbf{Decomposition of task-agnostic general knowledge.} We decompose the regularization term $\Lkl$ in Eq.~\eqref{eq:klloss} as $\Lkl=\Lce+(\Lkl-\Lce)$. The first term $\Lce$ aims to learn task-specific knowledge. The second term ($\Lkl-\Lce)$ is the key component to learn supplementary knowledge that is provided by the task-agnostic general knowledge but not included in the task-specific downstream knowledge. 
To better understand the contribution of the two knowledge, we calculate their gradients \textit{w.r.t} the logit $\mathbf{z}$ of the model $f$ on the class $t$ as:
\begin{small}
\begin{equation}\label{eq:kl_id}
     \nabla_{\zt} \Lce=f_t-\mathbf{y}_t,
\end{equation}
\end{small}

\begin{small}
\begin{equation}\label{eq:kl_ood}
    \nabla_{\zt} (\Lkl-\Lce)=(f_t-\yzs)-(f_t-\mathbf{y}_t)=\mathbf{y}_t-\yzs_t.
\end{equation}
\end{small}

We have the following observations: First, the two gradients always have opposite directions, learning one knowledge well will inevitably lead to the deterioration of the other knowledge. If $t$ is the true class, \ie, $\mathbf{y}_t=1$,  then $\nabla_{\zt} \Lce=f_t-1 < 0$ while $\nabla_{\zt} (\Lkl-\Lce)=1-\yzs_t > 0$. If $t$ is the false class, \ie, $\mathbf{y}_t=0$, $\nabla_{\zt} \Lce=f_t > 0$ while $\nabla_{\zt} (\Lkl-\Lce)=-\yzs_t < 0$. The first observation reveals the conflict between task-specific knowledge learning and supplementary knowledge learning, which motivates us to balance the learning process of the two knowledge.
     
Second, the gradient  $\mathbf{y}_t-\yzs_t$ for learning supplementary knowledge is constant in different optimization steps, and the magnitude varies sample-wisely. As a comparison, the gradient $f_t-\mathbf{y}_t$ for learning task-specific knowledge {keeps updated} during the training process ($f_t$ changes after each optimization step). The above analysis motivates us to trade-off the two knowledge in a sample-wise and dynamic manner, dubbed ProReg:

\begin{equation}\label{eq:totalloss0}
 \LProReg=\Lce+w\cdot (\Lkl-\Lce)= (1-w) \cdot \Lce + w\cdot \Lkl.
\end{equation}

The sample-wise trade-off weight $w$ aims to prevent the model from biasing to either task-specific knowledge and supplementary knowledge. There are two typical cases that affect the determination of $w$.

If $|\nabla_{\zt} \Lce| $ is much larger than $|\nabla_{\zt} (\Lkl-\Lce)|$, 
the task-specific knowledge will dominate the overall optimization direction of this sample. We assign a larger weight $w$ to the term $(\Lkl-\Lce)$ to emphasis the supplementary knowledge and prevent the model from biasing to the downstream knowledge. 
On the contrary, if $|\nabla_{\zt} (\Lkl-\Lce)|$ is much larger than $|\nabla_{\zt} \Lce|$, the task-specific downstream knowledge might be neglected. We assign a smaller weight $w$ to $(\Lkl-\Lce)$ to guarantee the learning of downstream knowledge.
From the above analysis, the trade-off weight $w$ should be proportional to 
$|{\nabla_{\zt} \Lce}|/|{\nabla_{\zt} (\Lkl-\Lce)}|$ to balance the two knowledge. From Eq.~\eqref{eq:kl_id}
 and Eq.~\eqref{eq:kl_ood}, we have $|{\nabla_{\zt} \Lce}|/|{\nabla_{\zt} (\Lkl-\Lce)|}=\frac{\mathbf{y}_t-f_t}{\mathbf{y}_t-\yzs_t}$. For simplifying analysis, we consider the binary classification case and assume the ground-truth label is $t$, \ie, $\mathbf{y}_t=1$. For true class $t$, we have $\frac{\mathbf{y}_t-f_t}{\mathbf{y}_t-\yzs_t}=\frac{1-f_t}{1-\yzs_t} \propto \frac{\yzs_t}{f_t}$. For the false class $(1-t)$, we have $\frac{\mathbf{y}_{1\!-\!t}-f_{1\!-\!t}}{\mathbf{y}_{1\!-\!t}-\yzs_{1\!-\!t}}=\frac{0-(1-f_t)}{0-(1-\yzs_t)} \propto \frac{\yzs_t}{f_t}$. 
 From Eq.~\eqref{eq:totalloss0}, we also expect $w\in [0,1]$ to guarantee the positive sign of $\Lce$ and $\Lkl$. Therefore, we design the trade-off weight as
 \begin{equation}
     w=\frac{f_t}{f_t+\yzs_t} \propto \frac{\yzs_t}{f_t},
 \end{equation}
 where $t$ is the ground-truth label.
 
 In our implementation, in addition to the sample-wisely trade-off weight $w$, we further introduce a hyper-parameter $\alpha>0$ on $\Lkl$ in Eq.~\eqref{eq:totalloss0} to control the strength of trade-off globally. Our ProReg is formulated as:
\begin{equation}\label{eq:totalloss_f}
\LProReg=(1-w) \cdot \Lce + \alpha \cdot w\cdot \Lkl
\end{equation}

%% file: tables/prompt_example.tex
\begin{table}[t]
\centering
\scalebox{0.9}{
\begin{tabular}{lll}
\toprule
\textbf{\begin{tabular}[c]{@{}l@{}}Downs.\\ Task\end{tabular}}        & \textbf{\begin{tabular}[c]{@{}l@{}}Pretext\\ Task\end{tabular}} &   \multicolumn{1}{c}{\textbf{Prompt Example}} \\ \midrule
\multirow{2}{*}{\begin{tabular}[c]{@{}l@{}}Image \\ CLS\end{tabular}} & \multirow{2}{*}{Contra.} &\textbullet\-  the person is \class.       \\
                                                                                 &                      &\textbullet\- a photo of a \class.         \\ \midrule
\multirow{8}{*}{VQA}  & \multirow{4}{*}{MLM} &  \begin{tabular}[c]{@{}l@{}} \textbullet\- how many hats are there?\\ \quad $\xrightarrow{\text{Q2S}}$ there are \mask hats.  \end{tabular}    \\  
                      &                      &  \begin{tabular}[c]{@{}l@{}} \textbullet\- what color is the shirt?\\  $\quad \xrightarrow{\text{Q2S}}$ the color of the shirt is \mask.\end{tabular}   \\ \cline{2-3} 
                      & \multirow{4}{*}{ITM} &\begin{tabular}[c]{@{}l@{}} \textbullet\- does this fruit grow on vines?\\ \quad $\xrightarrow{\text{Q2S}}$ this fruit grow on vines.  \end{tabular}      \\
                       &                     &\begin{tabular}[c]{@{}l@{}}\textbullet\- is the zebra sleeping? \\ \quad $\xrightarrow{\text{Q2S}}$ the zebra is sleeping. \end{tabular}   \\ \bottomrule
\end{tabular}
}
\caption{Prompt Examples. For VQA, the prompt is generated according to the question type. If the question is open-ended, we convert the question to a statement with masked token and use MLM head to predict the answer. If the question is closed-ended, we convert the question to a statement and use ITM head to predict ``yes'' or ``no''. For image classification, we adopt a simple static template and use cosine similarity to predict the class.}
\label{tab:promptexample}
\end{table}

%% file: sections/4-experiments.tex
\section{Experiments}

\subsection{Datasets and Implementation Details}
\input{images/dataset}
\noindent\textbf{BAR}~\cite{LfF} is a real-world image dataset for action classification, where the action is biased to the place. Examples of ``climbing'' class are shown in Figure~\ref{fig:dataset}(a), where most training background are rocks, while the one of testing images are snow.

\noindent\textbf{NICO}~\cite{nico} dataset is designed for Out-of-Distribution (OOD) image classification, which has 2 subsets (animal and vehicle), 19 classes and 188 contexts. For each class, we select 3 different contexts for both training and test set, \eg, the training contexts for ``sheep'' are ``in water'', ``on road'' and ``on snow'', while the test contexts are ``at sunset'', ``on grass'' and ``in forset''. Please kindly refer to Apprendix for more details of our setting.

\noindent\textbf{PACS}~\cite{PACS} covers photo, sketch, cartoon and painting domains. Figure~\ref{fig:dataset}(b) shows some examples. The model is trained and validated on any three seen domains then tested on the rest unseen domain.

\noindent\textbf{DomainNet}~\cite{DomainNet} consists of images from 345 classes covering the ``sktech'', ``real'', ``clipart'' and ``painting'' domains. Here, we use the ``sktech'' domain as ID dataset, and use ``clipart'' and ``painting'' as the OOD datasets. Please kindly refer to Appendix for more details.

\noindent\textbf{VQA-CP}~\cite{VQACP-GVQA} is proposed to examine the generalizability of VQA models when the language prior varies significantly between training and testing splits. Figure~\ref{fig:dataset}(c) shows some answer distribution of the training and test set, \eg, most of “what color ” questions are answered as “white” in training set while “black ” in the test set.

\noindent\textbf{Experimental Details.}
For ViLT-based models, we followed the original fine-tuning settings in \cite{vilt}, which adopt the ViLT-B/32 model with AdamW~\cite{AdamW} optimizer for $10$ epochs for all datasets. 
For CLIP-based models, we used the ViT-B/32 backbone and adopted the ViLT fine-tuning settings, including the training epoch, optimizer, warmup schedule and image pre-processing, \etc. $\alpha$ is set to 2 for all experiments.
 See Appendix for more details.

\noindent\textbf{Fine-tuning Baselines.} 
For CLIP-based models, we compared ProReg with 6 baselines.
(1) zero-shot CLIP~\cite{clip};
(2) linear probe~\cite{clip};
(3) prompt tuning, \ie, CoOp~\cite{coop};
(4) an advance CLIP fine-tuning: CLIP-adapter~\cite{gao2021clip};
(5) conventional fine-tuning with randomly initialized classification head, which is denoted as FT;
(6) conventional fine-tuning with classification head initialized from a text encoder thought prompting, which is denoted as FT++;

For ViLT-based models, we compared ViLT-ProReg with three baseline methods. 
(1) zero-shot ViLT. 
(2) conventional fine-tuning with randomly initialized classification head, which is denoted as FT. 
(3) conventional fine-tuning with classification head initialized from a text encoder thought prompting, which is denoted as FT++.
 Please see Appendix for more details.

\noindent\textbf{Evaluation Metrics.} To evaluate the unbiasedness, we also report the in-domain (ID) accuracy, out-of-domain (OOD) accuracy and their harmonic mean for NICO and DomainNet datasets. Specifically, ID test set has the same distribution with the training set while the distribution of OOD test set is different from the training one. A debiased model should have high performance on both ID and OOD testing, as a result, it should also have the highest harmonic mean. 

\subsection{Main Results}\label{sec:main_result}
\input{tables/BAR_DomainNet}
\input{tables/NICO}

\noindent\textbf{Image Classification.} Results are shown in Table~\ref{tab:bar_result}, Table~\ref{tab:nico_result}, Table~\ref{tab:pacs_result} and Table~\ref{tab:domain_net_result}.
Due to the powerful pretraining knowledge and hand-crafted prompt, the zero-shot CLIP model achieves strong performance.
In particular, on NICO Vehicle subset (Table~\ref{tab:nico_result}), the zero-shot CLIP exhibits harmonic mean of $95.8\%$. 
After training on the downstream data, we observed that ProReg demonstrates clear advantages over other fine-tuning methods on the OOD test sets, \eg, on BAR test set, the ProReg outperforms other counterparts by at least $1.2\%$. 
Not surprisingly, on NICO and DomainNet datasets, we observed that other fine-tuning methods gain significant improvements on in-distribution accuracies but at a cost of performance loss on out-of-distribution accuracies compared to zero shot performance, \eg, recently proposed CoOp~\cite{coop} increases the ID accuracies from $93.3\%$ to $99.3\%$ on NICO Animal subset (from $58.2\%$ to $70.9\%$ on DomainNet) but the OOD accuracies decreases from $92.5\%$ to $86.1\%$ (from $63.7\%$ to $56.6\%$ on DomainNet). 
As a comparison, our ProReg 
obtains a good trade-off between on ID and OOD performance, thus achieving the best performance on harmonic means.

Table~\ref{tab:pacs_result} reported the results on PACS. 
The zero-shot CLIP shows strong performance on Photo (P) and Cartoon (C) Domains, \eg, zero-shot CLIP achieves $99.4\%$ on photo domain.
Besides the best performance in average accuracy, our ProReg significantly outperforms other fine-tuning methods on difficult unseen domains Sketch (S) and Cartoon (C), \eg, ProReg gains $1.6\%$ and $2.6\%$ on sketch and cartoon domain compared to FT++ method. These results show the power of our ProReg to overcome diverse domain biases.  
\input{tables/pacs}

\input{tables/vqacp}
\noindent\textbf{Visual Question Answering.} Considering that the input text for VQA tasks varies sample-wisely, it is difficult for CLIP model to infer and optimize. We only implement the ProReg based on ViLT models and compared with zero-shot prompt, FT and FT++.

Although zero-shot prompt models have no access to the training data, thanks to the pretraining knowledge and our proposed question-to-statement (Q2S) prompt design, our zero-shot prompt achieves impressive results with $43.62\%$ accuracy, which is a strong baseline.
The results show that our ProReg framework shows strong performance under language bias, \eg, ProReg achieves an overall accuracy of $54.89\%$ with an impressive improvement $+8.55\%$ compared to conventional fine-tuning with classification initialized from prompt (FT++). More interestingly, conventional promptless fine-tune (FT) failed on ``Yes/No'' questions, while FT++ failed on ``Other''. As a comparison, ProReg performs relatively well on all the three questions types.

\subsection{Ablation Studies}\label{sec:abla}

\input{images/alpha}

\noindent\textit{Q1: What is the effect of the hyper-parameter $\alpha$?} We conducted the experiments on DomainNet dataset by varying $\alpha$ in Eq.~\eqref{eq:totalloss_f}, the results are shown in Figure~\ref{fig:alpha}. As expected, as $\alpha$ increases, the fine-tuned model is encouraged to learn more from pre-trained knowledge. As a result, the OOD performance will increase while the ID accuracy drops.  

\input{images/distill_main}
\noindent\textit{Q2: Can we blend the 
knowledge using conventional distillation that use constant trade-off weight?} No, its performance is worse than ProReg fine-tune.
We implemented traditional knowledge distillation strategy by varying the trade-off weight $\lambda$ as described in  Eq.~\eqref{eq:totalloss1}. Figure \ref{fig:distill_main} shows the results on NICO Vehicle and VQA-CP dataset. 
We observed that our ProReg model achieves superior performance than traditional knowledge distillation, \eg, for VQA-CP datasets, even with the optimal $\lambda$, traditional KD (blue line) achieves the best performance of $53.15\%$ with $-1.74\%$ gap to ProReg fine-tune (red line).

\input{images/ensemble_main}
\noindent\textit{Q3: Can we directly ensemble the zero-shot model and traditional fine-tune model?} 
No, it not only achieves worse performance than ProReg, but also doubles the inference time and the number of parameters. 
In Figure~\ref{fig:ensemble_main}, we investigate whether combining the knowledge by ensembling the conventional fine-tuning and the zero-shot CLIP model can perform superior results. Specifically, given the prediction of fine-tuning model $\mathbf{y}^\text{ft}$ and the one from zero-shot model $\yzs$, the ensembled prediction is formulated as
   $ \mathbf{y}^\text{ens} = (1-\lambda)\cdot \mathbf{y}^\text{ft} + \lambda \cdot \yzs,$
where $\lambda \in [0,1]$. Empirical results on VQA-CP dataset and NICO vehicle dataset are plotted in Figure~\ref{fig:ensemble_main}(a), where the highest harmonic mean wiht $\lambda=0.5$ (blue line) still underperforms our ProReg (red line). For VQA-CP dataset, ensemble model reaches its optimal accuracy $53.48\%$ with $\lambda=0.5$ (blue line). The highest accuracy of ensemble is still surpassed by our ProReg result ($54.89\%$, red line). Moreover, the ensemble model doubles the inference time and the number of parameters. 

\noindent\textbf{Qualitative results.} 
\input{images/failure_case_vqa}
We visualized a sports related question, which is too domain-specific, the general knowledge learned by zero-shot prompt model is hard to answer such question.
In the meanwhile, our ProReg inherits the knowledge from both domain and gives the right answer.
Figure~\ref{fig:failure_vqa}(a) and (b) show some failure cases of conventional fine-tune model and zero-shot ViLT model for BAR dataset. In Figure \ref{fig:failure_vqa} (a), a ``climb'' image is mis-classified as ``vault'' by conventional fine-tune model, which can be attributed to the dataset bias, where the context of the ``climb'' training images are most rocks.
In Figure \ref{fig:failure_vqa} (b), a ``vault'' image is recognized as ``dive'' by the zero-shot ViLT model, we conjecture that the reason is that the knowledge of pole vaulting is not common in the pretrained domain. In both cases, ProReg predicts right answers by inheriting knowledge from the downstream data and the pretraining knowledge.

%% file: images/dataset.tex
\begin{figure}[t]
    \centering
    \includegraphics[width=0.45\textwidth]{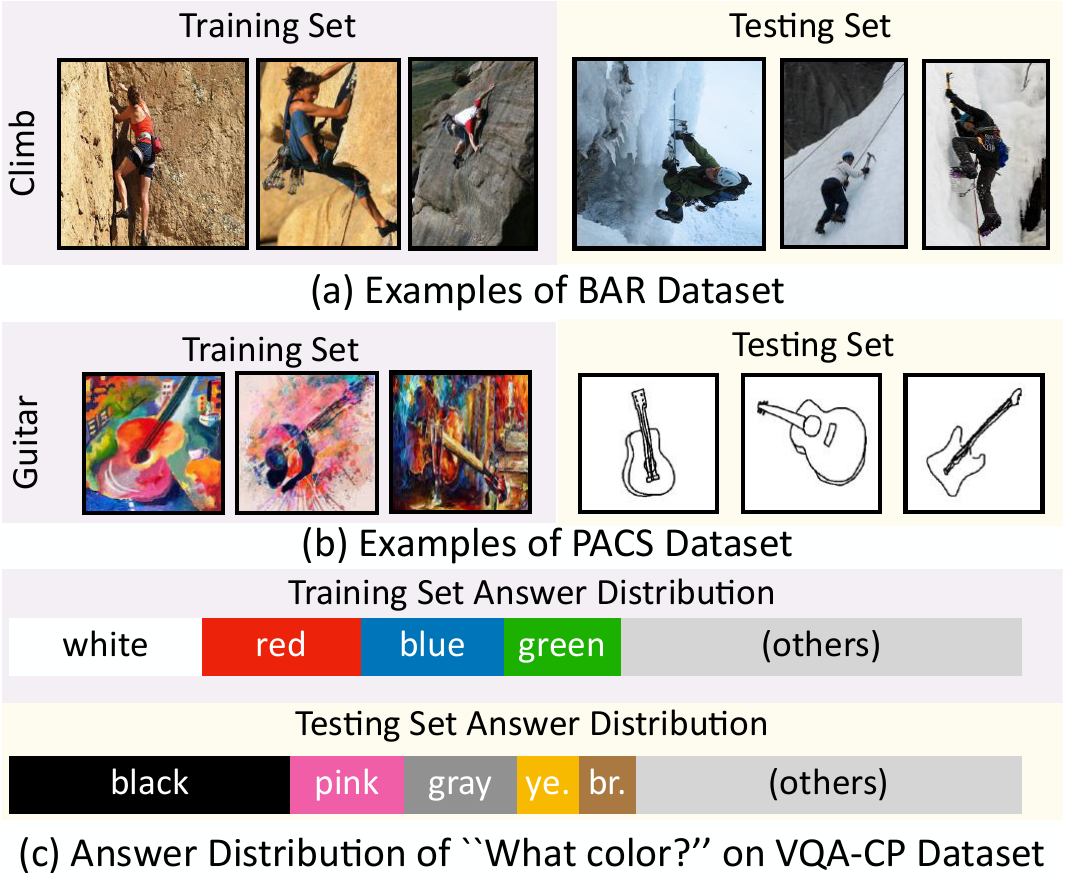}
    \caption{Examples of OOD benchmarks.}
    \label{fig:dataset}
\end{figure}

%% file: tables/BAR_DomainNet.tex
\begin{table}
\scalebox{0.9}{
\begin{minipage}[t]{0.2\textwidth}
\centering
\setlength{\tabcolsep}{4pt}
\begin{tabular}{ll}
\toprule
Method             & Acc. \\
\midrule
Zero-shot & 86.7   \\
Linear Probe & 89.5 \\
CoOp      & 90.2   \\
CLIP-Adapter & 90.4 \\
FT &  90.5 \\
FT++ & 90.9  \\
\rowcolor{tabhighlight}
\textbf{ProReg}    & \textbf{92.1}  \\
\bottomrule
\end{tabular}
\caption{Accuracy (\%) on BAR.}
\label{tab:bar_result}
\end{minipage}
\hspace{1mm}
\begin{minipage}[t]{0.3\textwidth}
\centering
\setlength{\tabcolsep}{4pt}
\begin{tabular}{llll}
\toprule
 Method                        & ID      & OOD      & HM               \\ \midrule
Zero-shot  &  58.2  & 63.7   & 60.8         \\
Linear Probe & 70.4    &   52.9&  60.4           \\
CoOp & 70.9  &     56.6 & 62.9 \\
CLIP-Adapter & 71.2  & 51.8 & 60.0\\
FT    & \bf 75.9  & 55.8 &  64.3 \\
FT++  & 75.8 & 60.6 & 67.4  \\
\rowcolor{tabhighlight}
\bf ProReg  & 75.6 & \bf 64.4 & \bf 69.2     \\
\bottomrule 
\end{tabular}
\caption{Accuracy (\%) on DomainNet dataset.}
\label{tab:domain_net_result}
\end{minipage}
}
\end{table}

%% file: tables/NICO.tex
\begin{table}[t]
\setlength{\tabcolsep}{4pt}
\label{table:nico}
\centering
\scalebox{0.9}{
\begin{tabular}{lcccccc}
\toprule
\multirow{2}{*}{Method} & \multicolumn{3}{c}{Animal Subset} & \multicolumn{3}{c}{Vehicle Subset} \\ \cline{2-7} 
                        & ID    & OOD        & HM       & ID        & OOD       & HM   \\ \midrule
Zero-shot   & 93.3  & 92.5       & 92.9     & 95.8      &  95.9     &  95.8      \\ 
Linear Probe& 99.0  & 87.0       & 92.6     & 99.4   &  87.5     &  93.1     \\
CoOp   &\bf 99.3    & 86.1       & 92.2  & 99.4   &  87.6     &  93.1     \\
CLIP-Adapter & 99.0 & 81.9 & 89.6 & 99.4 & 83.7& 90.9 \\
FT           & 98.7  & 85.3       & 91.5   &\bf 99.6   & 88.9      & 93.9    \\     
FT++         & 98.9  & 91.7       & 95.2     & 99.4   &  90.8     &  94.9     \\
\rowcolor{tabhighlight}
\textbf{ProReg}  & 98.5  & \bf 94.4   & \bf 96.4 & 98.7      & \bf 94.1  & \bf 96.3     \\
\bottomrule
\end{tabular}}
\caption{Accuracy (\%) on NICO dataset.}
\vspace{-4mm}
\label{tab:nico_result}
\end{table}

%% file: tables/pacs.tex
\begin{table}
\centering
\setlength{\tabcolsep}{4pt}
\scalebox{0.9}{
\begin{tabular}{llllll}
\toprule
\multirow{2}{*}{Method} & \multicolumn{4}{c}{OOD Domain}       & \multirow{2}{*}{Avg} \\ \cline{2-5}
                       & A      & C       & P     & S     &                   \\ \midrule
Zero-shot  & 89.8   & 96.3    & 99.4  & 85.8  &   92.8           \\
Linear Probe & 84.3   & 91.7    & 83.1  & 80.3  &  84.9         \\
CoOp & 92.6   & 96.1    & 98.1  & 85.0  &   93.0  \\
CLIP-Adapter & 93.6 & 96.6 & 97.7 & 84.4 & 93.1\\
FT   & 95.3   & 95.6    & 99.7  & 86.8  &  94.4 \\
FT++  & 95.6   & 97.0    & 99.3  & 86.8  &   94.7           \\
\rowcolor{tabhighlight}
\bf ProReg    &\bf 96.2& \bf 98.6&\bf 99.8 & \bf 89.4 & \bf 96.0            \\
\bottomrule 
\end{tabular}}
\caption{Accuracy (\%) on PACS.}
\label{tab:pacs_result}
\end{table}

%% file: tables/vqacp.tex
\begin{table}[t]
\centering
\setlength{\tabcolsep}{4pt}
\label{tab:vqacp_result}
\scalebox{0.9}{
\begin{tabular}{llQcccc}
\toprule
& Methods & ~ & {\bf All} & {Y/N} &  {Num.} & {Other} \\
\hline
\multirow{4}{*}{\rotatebox{90}{ViLT}}
& Q2S Zero-shot  & ~ & 43.62 & \textbf{77.70} & 12.02 & 33.93  \\
&FT   & ~ & 45.68 & 44.16 & 14.44 & 55.12 \\
&FT++   & ~ & 46.34 & 73.00 & \textbf{16.29}  & 40.23  \\
\rowcolor{tabhighlight}
&\bf ProReg & ~ & \textbf{54.89} & 73.06 & 14.85  & \bf 55.58     \\
\bottomrule
\end{tabular}}
\caption{Evaluations (Accurarcy\%) on VQA-CP.}
\end{table}

%% file: images/alpha.tex
\begin{figure}[t]
    \centering
    \includegraphics[width=0.45\textwidth]{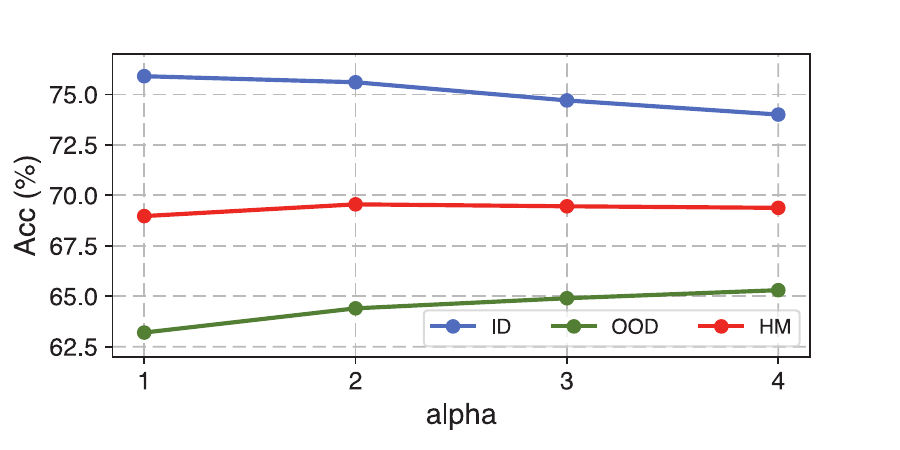}
    \caption{Effect of hyper-parameter $\alpha$ on DomainNet.}
    \label{fig:alpha}
\end{figure}

%% file: images/distill_main.tex
\begin{figure}[t]
\centering
\includegraphics[width=0.45\textwidth]{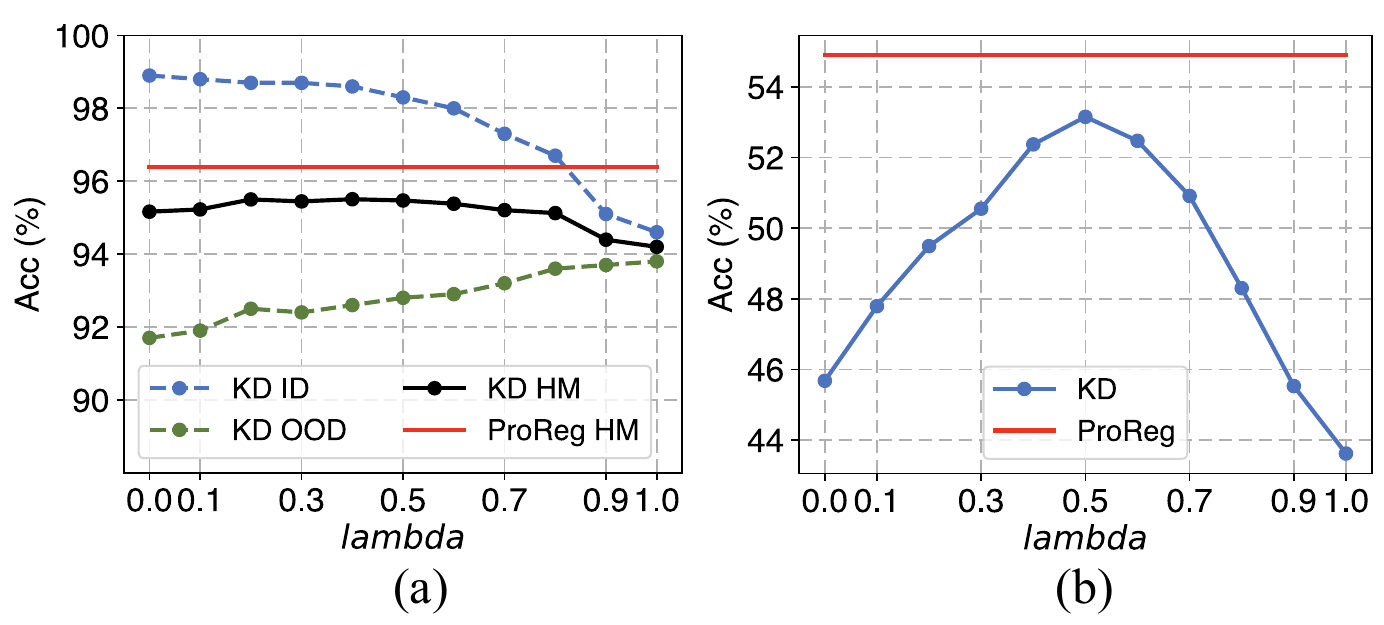}
\caption{(a) KD results (ID, OOD and harmonic $\%$) on NICO animal subset.(b) KD results (overall accuracy $\%$) on VQA-CP with different weight $\lambda$.}
\label{fig:distill_main}
\end{figure}

%% file: images/ensemble_main.tex
\begin{figure}[t]
\centering
\includegraphics[width=0.45\textwidth]{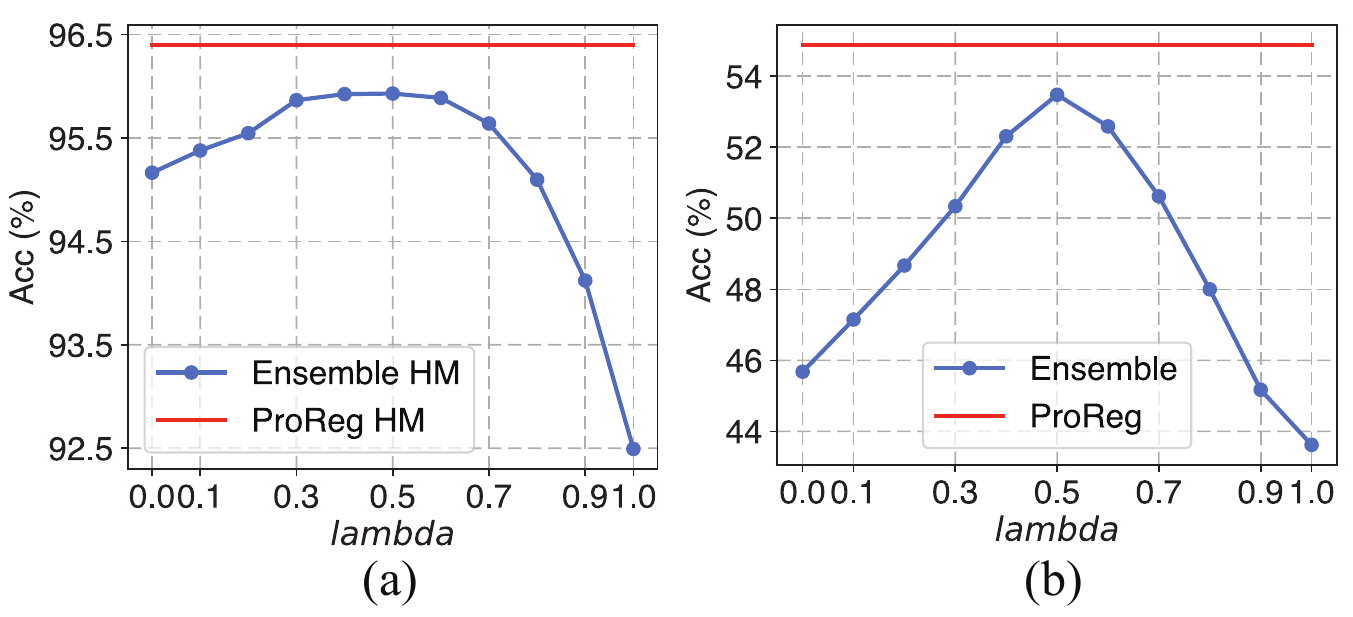}
\caption{(a) Ensemble results (harmonic mean $\%$) on NICO vehicle subset. (b) Ensemble results (overall accuracy $\%$) on VQA-CP with different weight $\lambda$.}
\label{fig:ensemble_main}
\end{figure}

%% file: images/failure_case_vqa.tex
\begin{figure}[t]
\centering
\includegraphics[width=0.45\textwidth]{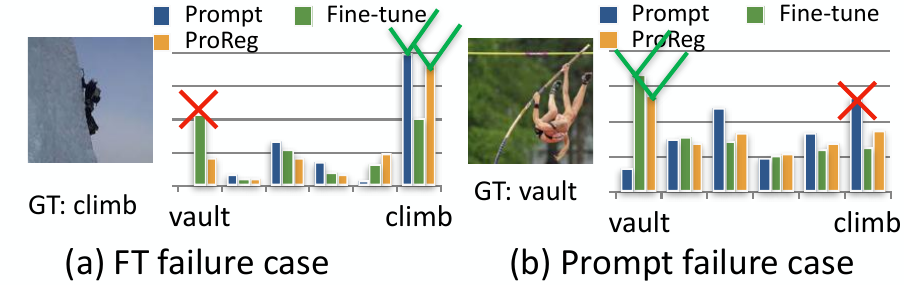}
   \caption{Qualitative examples of zero-shot prompt, conventional fine-tune and ProReg fine-tune on BAR dataset.}
   \label{fig:failure_vqa}
\end{figure}

%% file: sections/5-conclusion.tex
\section{Conclusion}
We presented an effective novel fine-tuning strategy: Prompt Regularization (ProReg), for deploying VLMs in downstream tasks. As its name implies, ProReg is motivated to make the best of the two worlds: pretrained general knowledge and task-specific knowledge, where the former is acquired by using prompt. The highlight of ProReg is the proposed sample-wise adaptive weight that trades off the training losses from the two worlds. Note that this weight is theoretically justified. Therefore, we believe that ProReg has a great potential for helping practitioners to fine-tune their own models efficiently. By extensive evaluations on three types of challenging OOD benchmarks, ProReg significantly outperformed zero-shot prompt, prompt tuning, conventional fine-tuning and other state-of-the-art methods. 

%% file: sections/6-ack.tex
\section*{Acknowledgments}
The authors would like to thank the reviewers for their comments that help improve the manuscript. 
This research is supported by the National Research Foundation, Singapore
under its AI Singapore Programme (AISG Award No: AISG2-PhD-2021-01-002 and AISG2-RP-2021-022) and MOE Tier2 MOE2019-T2-2-062.